\documentclass[10pt,onecolumn,letterpaper]{article}

\usepackage{cvpr}
\usepackage{times}
\usepackage{epsfig}
\usepackage{graphicx}
\usepackage{amsmath}
\usepackage{commath}
\usepackage{amssymb}
\usepackage{enumerate,booktabs}
\usepackage{subfigure,color}
\usepackage[numbers,sort&compress]{natbib}
\usepackage{multirow}

\usepackage[pagebackref=true,breaklinks=true,letterpaper=true,colorlinks,bookmarks=false]{hyperref}

\cvprfinalcopy 

\def\cvprPaperID{XXXX} 

\begin{document}
\def\mA{\mathcal{A}}
\def\mB{\mathcal{B}}
\def\mC{\mathcal{C}}
\def\mD{\mathcal{D}}
\def\mE{\mathcal{E}}
\def\mF{\mathcal{F}}
\def\mG{\mathcal{G}}
\def\mH{\mathcal{H}}
\def\mI{\mathcal{I}}
\def\mJ{\mathcal{J}}
\def\mK{\mathcal{K}}
\def\mL{\mathcal{L}}
\def\mM{\mathcal{M}}
\def\mN{\mathcal{N}}
\def\mO{\mathcal{O}}
\def\mP{\mathcal{P}}
\def\mQ{\mathcal{Q}}
\def\mR{\mathcal{R}}
\def\mS{\mathcal{S}}
\def\mT{\mathcal{T}}
\def\mU{\mathcal{U}}
\def\mV{\mathcal{V}}
\def\mW{\mathcal{W}}
\def\mX{\mathcal{X}}
\def\mY{\mathcal{Y}}
\def\mZ{\mathcal{Z}}

\def\1n{\mathbf{1}_n}
\def\0{\mathbf{0}}
\def\1{\mathbf{1}}

\def\A{{\bf A}}
\def\B{{\bf B}}
\def\C{{\bf C}}
\def\D{{\bf D}}
\def\E{{\bf E}}
\def\F{{\bf F}}
\def\G{{\bf G}}
\def\H{{\bf H}}
\def\I{{\bf I}}
\def\J{{\bf J}}
\def\K{{\bf K}}
\def\L{{\bf L}}
\def\M{{\bf M}}
\def\N{{\bf N}}
\def\O{{\bf O}}
\def\P{{\bf P}}
\def\Q{{\bf Q}}
\def\R{{\bf R}}
\def\S{{\bf S}}
\def\T{{\bf T}}
\def\U{{\bf U}}
\def\V{{\bf V}}
\def\W{{\bf W}}
\def\X{{\bf X}}
\def\Y{{\bf Y}}
\def\Z{{\bf Z}}

\def\a{{\bf a}}
\def\b{{\bf b}}
\def\c{{\bf c}}
\def\d{{\bf d}}
\def\e{{\bf e}}
\def\f{{\bf f}}
\def\g{{\bf g}}
\def\h{{\bf h}}
\def\i{{\bf i}}
\def\j{{\bf j}}
\def\k{{\bf k}}
\def\l{{\bf l}}
\def\m{{\bf m}}
\def\n{{\bf n}}
\def\o{{\bf o}}
\def\p{{\bf p}}
\def\q{{\bf q}}
\def\r{{\bf r}}
\def\s{{\bf s}}
\def\t{{\bf t}}
\def\u{{\bf u}}
\def\v{{\bf v}}
\def\w{{\bf w}}
\def\x{{\bf x}}
\def\y{{\bf y}}
\def\z{{\bf z}}

\def\balpha{\mbox{\boldmath{$\alpha$}}}
\def\bbeta{\mbox{\boldmath{$\beta$}}}
\def\bdelta{\mbox{\boldmath{$\delta$}}}
\def\bgamma{\mbox{\boldmath{$\gamma$}}}
\def\blambda{\mbox{\boldmath{$\lambda$}}}
\def\bsigma{\mbox{\boldmath{$\sigma$}}}
\def\btheta{\mbox{\boldmath{$\theta$}}}
\def\bomega{\mbox{\boldmath{$\omega$}}}
\def\bxi{\mbox{\boldmath{$\xi$}}}
\def\bnu{\mbox{\boldmath{$\nu$}}}                                  
\def\bphi{\mbox{\boldmath{$\phi$}}}

\def\bDelta{\mbox{\boldmath{$\Delta$}}}
\def\bOmega{\mbox{\boldmath{$\Omega$}}}
\def\bPhi{\mbox{\boldmath{$\Phi$}}}
\def\bLambda{\mbox{\boldmath{$\Lambda$}}}
\def\bSigma{\mbox{\boldmath{$\Sigma$}}}
\def\bGamma{\mbox{\boldmath{$\Gamma$}}}

\newcommand{\myminimum}[1]{\mathop{\textrm{minimum}}_{#1}}
\newcommand{\mymaximum}[1]{\mathop{\textrm{maximum}}_{#1}}    
\newcommand{\mymin}[1]{\mathop{\textrm{minimize}}_{#1}}
\newcommand{\mymax}[1]{\mathop{\textrm{maximize}}_{#1}}
\newcommand{\mymins}[1]{\mathop{\textrm{min.}}_{#1}}
\newcommand{\mymaxs}[1]{\mathop{\textrm{max.}}_{#1}}  
\newcommand{\myargmin}[1]{\mathop{\textrm{argmin}}_{#1}} 
\newcommand{\myargmax}[1]{\mathop{\textrm{argmax}}_{#1}} 
\newcommand{\myst}{\textrm{s.t. }}

\newcommand{\denselist}{\itemsep -1pt}
\newcommand{\sparselist}{\itemsep 1pt}

\definecolor{pink}{rgb}{0.9,0.5,0.5}
\definecolor{purple}{rgb}{0.5, 0.4, 0.8}   
\definecolor{gray}{rgb}{0.3, 0.3, 0.3}
\definecolor{mygreen}{rgb}{0.2, 0.6, 0.2}

\newcommand{\cyan}[1]{\textcolor{cyan}{#1}}
\newcommand{\red}[1]{\textcolor{red}{#1}}  
\newcommand{\blue}[1]{\textcolor{blue}{#1}}
\newcommand{\magenta}[1]{\textcolor{magenta}{#1}}
\newcommand{\pink}[1]{\textcolor{pink}{#1}}
\newcommand{\green}[1]{\textcolor{green}{#1}} 
\newcommand{\gray}[1]{\textcolor{gray}{#1}}    
\newcommand{\mygreen}[1]{\textcolor{mygreen}{#1}}    
\newcommand{\purple}[1]{\textcolor{purple}{#1}}       

\definecolor{greena}{rgb}{0.4, 0.5, 0.1}
\newcommand{\greena}[1]{\textcolor{greena}{#1}}

\definecolor{bluea}{rgb}{0, 0.4, 0.6}
\newcommand{\bluea}[1]{\textcolor{bluea}{#1}}
\definecolor{reda}{rgb}{0.6, 0.2, 0.1}
\newcommand{\reda}[1]{\textcolor{reda}{#1}}

\def\changemargin#1#2{\list{}{\rightmargin#2\leftmargin#1}\item[]}
\let\endchangemargin=\endlist

\newcommand{\mtodo}[1]   {{\color{red}$\blacksquare$\textbf{[TODO: #1]}}}

\newcommand{\cm}[1]{}

\newcommand{\myheading}[1]{\vspace{0.05in}\noindent\textbf{#1}}

\newcommand\blfootnote[1]{%
  \begingroup
  \renewcommand\thefootnote{}\footnote{#1}%
  \addtocounter{footnote}{-1}%
  \endgroup
}

\title{Eigen Evolution Pooling for Human Action Recognition}

\author{Yang Wang$^\dagger$, Vinh Tran$^\dagger$, Minh Hoai\\
Stony Brook University, Stony Brook, NY 11794, USA \\
\{wang33, tquangvinh, minhhoai\}@cs.stonybrook.edu
}

\maketitle

\blfootnote{$^\dagger$ indicates equal contribution.}

\begin{abstract}
We introduce  Eigen Evolution Pooling, an efficient method  to aggregate a sequence of feature vectors. Eigen evolution  pooling is designed to produce compact feature representations for a sequence of feature  vectors, while maximally preserving as much information about the sequence as possible, especially the temporal evolution of the features over time.
Eigen evolution pooling is a general pooling method that  can be applied to any sequence of feature vectors, from  low-level RGB values to high-level Convolutional Neural Network (CNN) feature vectors. 
We show that  eigen evolution pooling is more effective than average, max, and rank pooling for encoding the dynamics of human actions in video. We demonstrate the power of eigen evolution pooling on UCF101 and Hollywood2 datasets, two human action recognition benchmarks, and achieve state-of-the-art performance.
\end{abstract}

\section{Introduction}
\label{sec:intro}

Human action recognition in video is a challenging problem because it is unclear how a video can be optimally represented. A current popular approach is to compute CNN features at multiple temporal locations of a video and subsequently use either average or max pooling to aggregate the feature vectors~\cite{Simonyan-Zisserman-NIPS14,wang2016temporal,Zhang-et-al-CVPR16,feichtenhofer2016convolutional,Tran-et-al-ICCV15}. This approach, however, fails to encode the long term dynamics of human actions exhibited in the sequence of feature vectors (or video frames). A better approach is to use rank pooling~\cite{Fernando-etal-CVPR15}, a recently proposed method that is specifically designed to capture the progression of  feature vectors in a sequence. Rank pooling has been shown to yield promising results, 
outperforming average and max pooling. However, rank pooling only encodes the overall trend  of the feature vectors; much information about the evolution of the feature vectors is not preserved. 

\begin{figure*}[t]
\begin{center}
\includegraphics[width=0.95\linewidth]{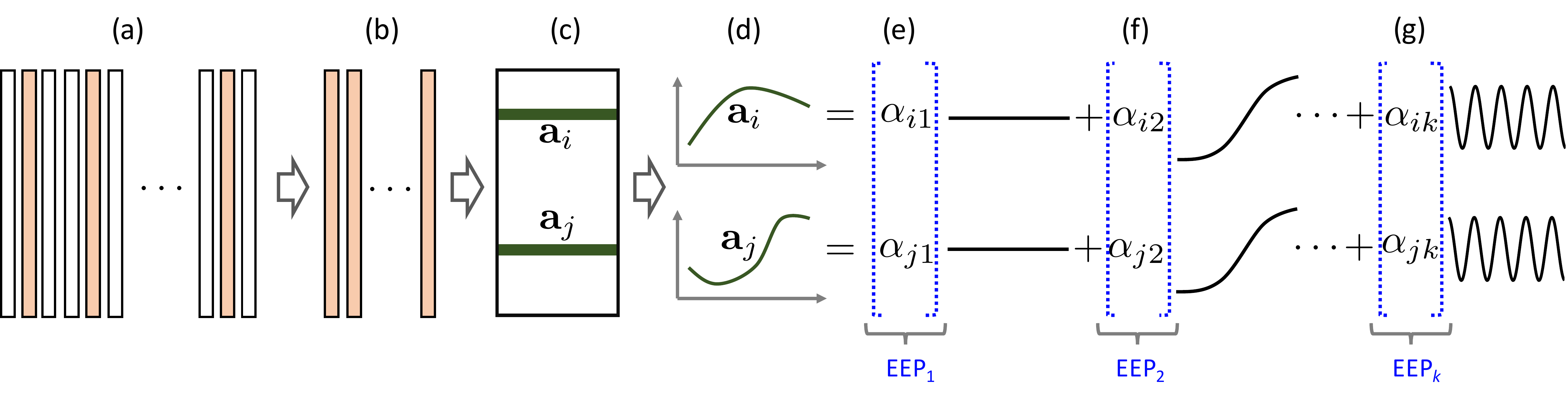}
\end{center}
\vskip -0.2in
   \caption{{\bf Eigen evolution pooling.}   (a): a sequence of feature vectors that need to be summarized. (b): $L$ vectors are sampled at a regular interval; $L$ is typically 16 or 25. (c, d): the sequence of sampled feature vectors can be viewed as an ordered set of one-dimensional functions. Each function can be decomposed as a linear combination of basis functions. The set of coefficients that correspond to a basis function defines an EEP pooling vectors, which will be referred to as $EEP_1, EEP_2, \cdots$, based on the order of the basis functions (e, f, g).}
\label{fig:eep}
\end{figure*}

In this paper, we propose Eigen Evolution Pooling (EEP), a temporal pooling method that 
preserves more information than rank pooling. The idea is to view a sequence of feature vectors as an ordered set of one-dimensional functions. Each function corresponds to the evolution of a feature over time, and the function can be expressed as a linear combination of basis functions. The basis functions can be optimally determined using Principle Component Analysis (PCA) to find the principle directions of feature evolution. Finally, the sequence of feature vectors is represented as one or several vectors of PCA coefficients. We refer to this process as Eigen Evolution Pooling, which is illustrated in Figure~\ref{fig:eep}.

Eigen evolution pooling provides an efficient way to summarize a video sequence into a compact and powerful representation. The eigen basis functions can be computed once and then used for encoding sequential data by a single matrix vector multiplication.  As will be explained in the next section, the eigen basis functions can be optimally computed by performing PCA on a collection of training data, but they can also be approximated using the basis functions of Discrete Cosine Transform.

Eigen evolution pooling can be applied to any sequence of feature vectors, from low-level features such as RGB values to high-level features from a CNN. When eigen evolution pooling is used to aggregate a sequence of RGB  frames, we will obtain a set of 3-channel images, which will be called eigen images.  Eigen images are analogous to the  recently proposed dynamic images~\cite{bilen2016dynamic}, which are obtained using rank pooling~\cite{Fernando-etal-CVPR15}. Alternatively, we can perform eigen evolution pooling on deep-learning features, e.g., TSN features~\cite{wang2016temporal}. 

Empirical evidence shows that eigen evolution pooling can encode the dynamics of humans actions in video better than average, max, and rank pooling methods.  As will also be seen in the experiments, the use of eigen evolution pooling on TSN features together with Dense Trajectory Descriptors or Video Darwin leads to a method that outperforms the current  state-of-the-art results on UCF101 and Hollywood2 datasets. More precisely, this method achieves the accuracy of 95.8\% on the UCF101 dataset~\cite{Soomro-et-al-TR12} and the mean average precision of 80.5\% on the Hollywood2~\cite{Marszalek-et-al-CVPR09} dataset.

\section{Related Works} \label{sec:related}

We propose a method for temporal pooling, encoding the temporal evolution of features over time. This is different from many existing action recognition methods that use orderless pooling such as Bag-of-Words \cite{Wang-et-al-BMVC09,Ullah-et-al-BMVC10,Wang-Schmid-ICCV13,peng2016bag,peng2013exploring,Kovashka-Grauman-CVPR10}, VLAD \cite{Jain-et-al-CVPR13}, Fisher Vector \cite{Wang-Schmid-ICCV13,wang2015action,Wang-Hoai-CVPR16}, and sparse coding \cite{Yang-et-al-CVPR09}. These methods aggregate hand-craft features such as SIFT3D \cite{Scovanner-et-al-MM07}, HOG3D \cite{Klaser-et-al-BMVC08}, DTD \cite{Wang-Schmid-ICCV13}, and TDD \cite{wang2015action} without taking into account their temporal order. Techniques such as pyramid pooling can be used to aggregate local features from each spatio-temporal grid cell, e.g.,~\cite{Laptev-et-al-CVPR08,Gaidon-et-al-BMVC12}. Such techniques, however, still fail to capture the long term dynamics of human actions.

Many state-of-the-art human action recognition methods~\cite{Simonyan-Zisserman-NIPS14,wang2016temporal,Zhang-et-al-CVPR16,feichtenhofer2016convolutional,Tran-et-al-ICCV15} are based CNN features.  Because CNNs are typically designed for  images, temporal pooling is needed for aggregating the extracted features from individual frames~\cite{Simonyan-Zisserman-NIPS14,wang2016temporal,Zhang-et-al-CVPR16,feichtenhofer2016convolutional} or blocks of frames \cite{Tran-et-al-ICCV15} selected at multiple temporal locations of a video. Many methods~\cite{Simonyan-Zisserman-NIPS14,wang2016temporal,Zhang-et-al-CVPR16,feichtenhofer2016convolutional,Tran-et-al-ICCV15} simply use average or max pooling, but this approach ignores the temporal evolution of the features.  An alternative solution is to use an LSTM recurrent neural network~\cite{Hochreiter-Schmidhuber-NC97} for temporal modeling, e.g.,~\cite{du2015hierarchical,donahue2015long,lev2015rnn}. However, these methods are computationally inefficient and require much training data.

\section{Eigen Evolution Pooling} \label{sec:eigen-evol-pooling} 

\subsection{Formulation}

Eigen evolution pooling is a general temporal pooling method that can be applied to any sequence of feature vectors  to encode the evolution of the features over time. The feature vectors must have the same dimensionality, but the vector sequence can have any length.  For eigen evolution pooling, we first sample $L$ vectors at a regular interval; typically, $L$ is 16 or 25. Let $\F = [\f_1, \cdots, \f_L] \in \mathbb{R}^{d\times L}$ represent a sequence of sampled feature vectors. Instead of considering $\F$ as a collection of columns, we propose to view $\F$ as a list of rows. Let $\a_i$ denotes the $i^{th}$ row of $\F$, i.e., $\F = [\a_1, \cdots, \a_d]^T$. Each row  $\a_i \in \mathbb{R}^L$ is a one-dimensional function that corresponds to the evolution of a feature over time. Instead of using the average value to summarize a function, we propose to represent it as a linear combination of basis functions, as illustrated in Figure~\ref{fig:eep}.

We propose to use a set of basis functions to  preserve as much information as possible. That corresponds to find the basis functions to minimize the reconstruction errors. Suppose we have a set of orthonormal basis functions $\G = [\g_1, \cdots, \g_k] \in \mathbb{R}^{L\times k}, \G^T\G=\I_k$. A function $\a$ can be decomposed into a linear combination of basis functions $\a \approx \G\c, \c \in \mathbb{R}^k$, and the coefficient vector $\c$ can be obtained as the product between the input function and the transpose of $\G$, i.e., $\c=\G^{T}\a$. Note that if $k$ (the number of basis functions) is small, the reconstructed function $\G\c=\G\G^T\a$ might not be exactly the same as the input function $\a$. In order to keep as much information as possible, we propose to find the optimal set of basis functions $\G$, by minimizing the reconstruction error:
\begin{align}
\label{eq: reconstruct-loss}
\G^* &= \myargmin{\G^T\G=\I_k}{\sum_{\F} \sum_{i}\lVert \G\G^T\a_i-\a_i \rVert^2}.
\end{align}

In the above, the first summation $\sum_{\F}$ refers to the enumeration over multiple video clips; each video clip leads to a sequence of sampled feature matrix $\F$. The second summation $\sum_{i}$ enumerates through the row of $\F$. Eq. (\ref{eq: reconstruct-loss}) is  equivalent to:
\begin{align}
\label{eq: simplified-loss}
\G^* = \myargmax{\G^T\G=\I_k}{ \sum_{j=1}^{k} \g_j^T \C \g_j},  \textrm{where } \C  = \sum_{\F} \sum_{i}{\a_i\a_i^T} = \sum_{\F} \F^T\F.
\end{align}

Matrix $\C$ is a covariance matrix. It is the covariance matrix between time steps, not the covariance matrix between features. The optimal set of basis functions $\G^*$ can be found using eigen decomposition: 
\begin{align}
\C=\sum_{i=1}^{L}{\lambda_i \e_i \e_i^T}, ~~\lambda_1 \ge \cdots \ge \lambda_L,
\end{align}
where $\e_1, \cdots, \e_L$ are the eigen vectors with corresponding eigen values $\lambda_1, \cdots, \lambda_L$. Since $\C$ is the covariance of features over times, we refer to $\e_1, \cdots, \e_L$ as \textsl{eigen evolution functions} or simply \textsl{eigen evolutions}. For smallest possible reconstruction error, we must have $\g_1 = \e_1,  \cdots, \g_k = \e_k$. 

For a basis function $\g$ and a feature sequence $\F$, $\F\g$ is the vector of coefficients corresponding to the basis function $\g$. This has the same dimension as the feature vectors in $\F$ and we refer to it as an Eigen Evolution Pooling (EEP) vector because $\g$ is an eigen evolution function. With different basis functions $\g$'s, we capture behaviors at different evolution directions. When $\e_1, \e_2, \cdots, \e_k$ are used as the basis function, we obtain $k$ different descriptors, which will be referred to as EEP$_1$, EEP$_2$, ..., EEP$_k$ respectively.

\subsection{Eigen Images -- eigen evolution pooling of RGB values} \label{sec:eigen-image}

Eigen evolution pooling is a general temporal pooling method that can be applied to any sequence of feature vectors. When it is applied directly to the RGB values of video frames, we will obtain eigen images, a simple yet effective representation that can summarize both the appearance and the dynamics of a video clip. An immediate benefit of eigen images is that they can be readily processed using some very successful and popular CNN architectures for action recognition. In this section, we will describe the process of constructing eigen images, which also illustrates how eigen evolution pooling works in general.

\begin{figure*}[t]
\begin{center}
\includegraphics[width=0.95\linewidth]{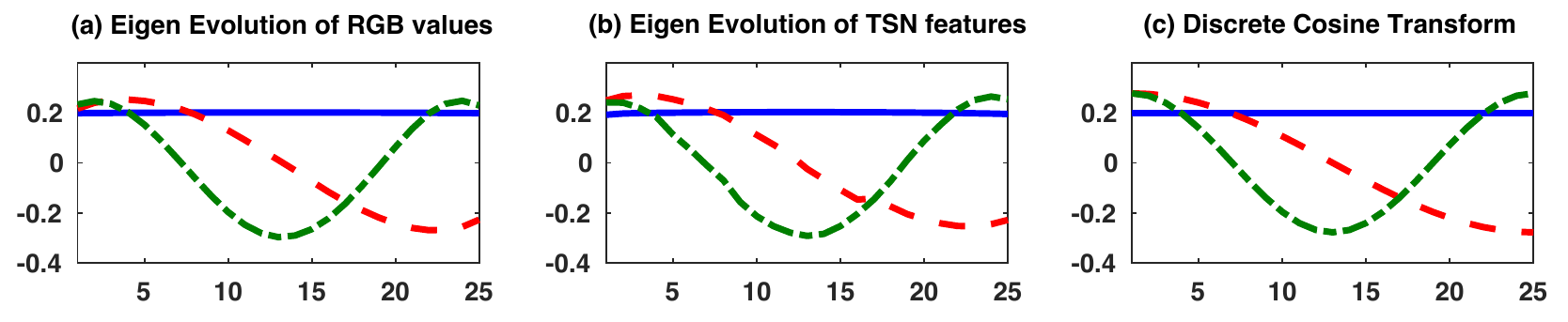}
\end{center}
\vskip -0.2in
   \caption{{\bf Exact and approximate eigen evolution functions.} (a) exact eigen evolution functions learned for sequences of RGB values. (b) exact eigen evolution functions for deep-learning TSN features. (c) approximate eigen evolution functions using the the basis functions of Discrete Cosine Transform (DCT). Interestingly, the eigen evolution functions for different types of feature vectors are similar, and they can be approximated by the basis functions of DCT.}
\label{fig:eigen_evolution}
\end{figure*}

Before constructing eigen images, we first need to compute the eigen evolution functions for the RGB values of individual pixels. The process is as follows. For each video clip, we evenly sample 25 temporal locations and obtain corresponding RGB images of size $256\times 340\times 3$. The images are vectorized and each video is represented as a sequence of vectorized RGB images $\textbf{F} \in \mathbb{R}^{261120\times 25}$. We subsequently compute $\textbf{C} = \sum_{\textbf{F}}{\textbf{F}^T\textbf{F}} \in \mathbb{R}^{25\times 25}$ as the covariance matrix between time steps, and perform eigen decomposition to obtain $\textbf{e}_1, \cdots, \textbf{e}_L$ as the eigen evolution functions. The first three eigen evolution functions for the RGB sequences are shown in Figure \ref{fig:eigen_evolution} (a).

After obtaining the eigen evolution functions, we can apply them as temporal pooling weights to efficiently compute the eigen images.  
Given a specific eigen evolution function $\textbf{g} =[ \alpha_{1},\cdots,\alpha_{L} ]^T$, for a video represented as a sequence of vectorized images $\textbf{F}=[\textbf{f}_1,\cdots,\textbf{f}_L]$, we can compute the corresponding eigen image as  $\sum_{l=1}^{L} \alpha_{l}\textbf{f}_l$.
The resulting image is reshaped to the original size of $256\times 340\times 3$, and the pixel values are rescaled to the range of $[0,255]$. 

Notably, eigen evolution pooling and rank pooling are both linear operators. Applying rank pooling to RGB images lead to the so-called dynamic image~\cite{bilen2016dynamic}. The pipeline to compute a dynamic image is similar to the pipeline to compute an eigen image, except the weight vector $\textbf{g}$ is the rank pooling weights: $\alpha_l = \sum_{t=l}^{L}{\frac{2t-L-1}{t}}$. 

We can also compute the eigen images and dynamic images locally within a sliding window, instead of computing them globally for the entire video. We set the window length to 16 frames, similar to Dense Trajectories~\cite{Wang-Schmid-ICCV13} and C3D~\cite{Tran-et-al-ICCV15}. With locally computed eigen images, we can capture the video dynamics at a finer scale.

\begin{figure*}[t]
\begin{center}
\includegraphics[width=0.95\linewidth]{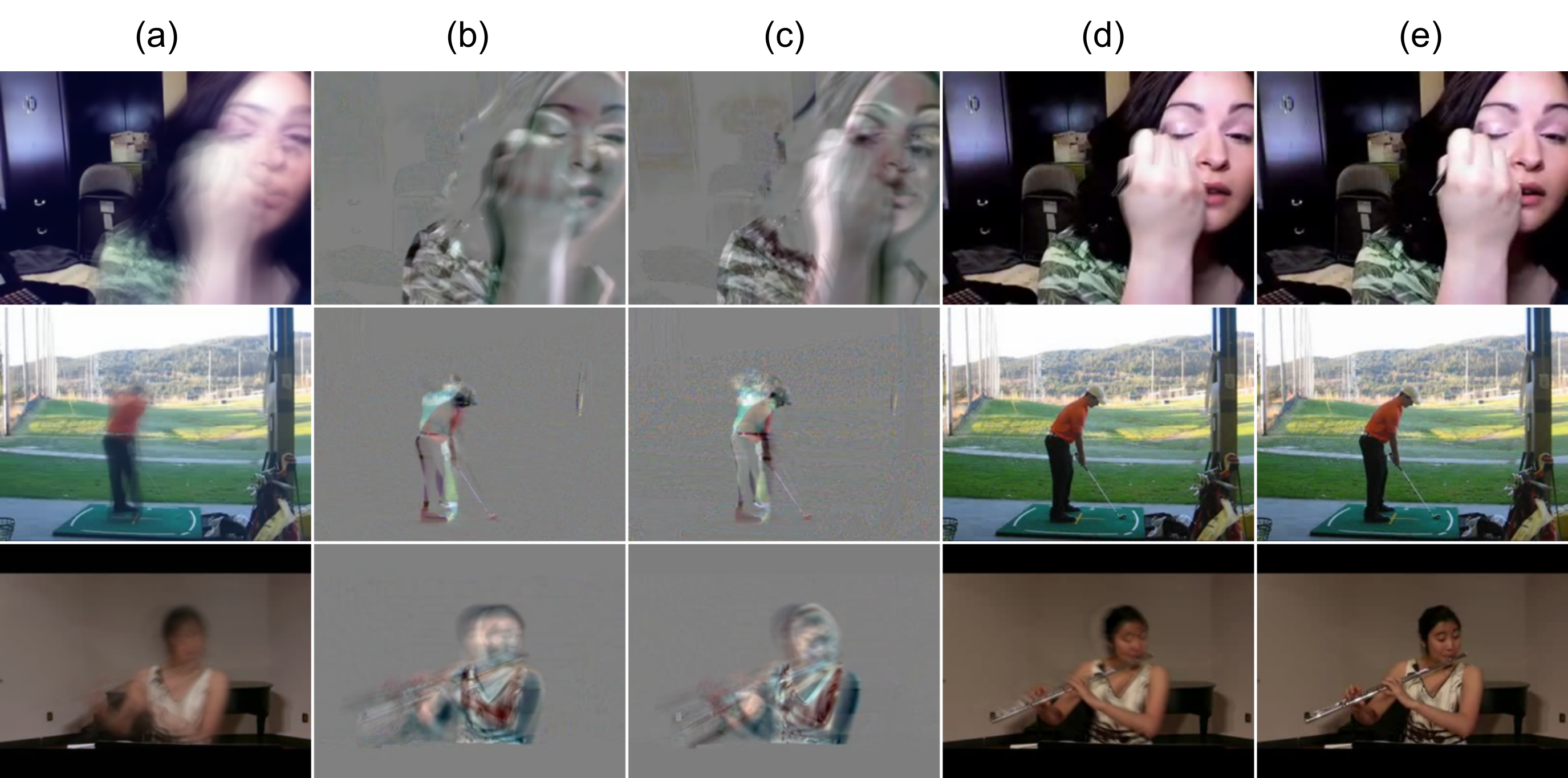}
\end{center}
\vskip -0.2in
   \caption{{\bf Eigen Images and reconstructed RGB frame.}  (a, b, c): eigen images computed with the first three eigen evolution functions. (d): the reconstructed image for the video frame at the middle of the sequence. (e): the original video frame at the middle of the video sequence. The reconstructed images are similar to the original images, indicating the sufficiency of using only three eigen images. Best viewed on a digital device.}
\label{fig:eigen_imgs}
\end{figure*}

\subsection{Eigen Evolution Functions \& Discrete Cosine Transform}

Figure \ref{fig:eigen_evolution} (a-b) displays the eigen evolution functions learned for two types of features: (a) low-level RGB values and (b) high-level deep-learning features. As can be observed, the learned eigen evolution functions  are similar for both feature types, and they can be approximated by the basis functions used in Discrete Cosine Transform (DCT), plotted in Figure \ref{fig:eigen_evolution}(c).
More specifically, the first eigen evolution is similar to an average function with a slight upward curve in the middle. The second and third eigen functions are similar to cosine functions oscillating at different frequencies.

The proposed eigen evolution pooling have the same formulation with Karhunen-Loeve Transform (KLT), except that KLT computes the basis functions using the correlation matrix instead of the covariance matrix between time steps. For strongly correlated Markov processes, the optimality of the eigen evolution pooling (or KLT) can be approached by DCT~\cite{ahmed2012orthogonal}. One advantage of DCT over eigen evolution pooling is that, DCT can be applied to sequential data of any duration. 
Thus we propose to use DCT basis functions to approximate eigen evolution functions, especially when we have to deal with feature vector sequences of different lengths and temporal subsampling leads to undesirable artifacts.

\section{Experiments} \label{sec:exp}

\subsection{Datasets}

We evaluate the proposed eigen evolution pooling on two public benchmarks: UCF101~\cite{Soomro-et-al-TR12} and Hollywood2~\cite{Marszalek-et-al-CVPR09}. The UCF101 dataset comprises 13,320 realistic action clips from 101 categories. Each category has at least 100 video clips, which were collected from YouTube. Each video contains a single action such as ``Archery'' or ``Basketball Dunk''. The dataset has three different training/test splits. We use top-1 accuracy as the evaluation metric for each training/test split in UCF101. The Hollywood2 dataset has 12 action classes with 1707 video clips collected from 69 different Hollywood movies. There are 823 videos in training set and 884 videos in testing set. We augment the training set with horizontally flipped training videos. We use mean Average Precision as the performance measurement.

\subsection{Eigen Images versus Dynamic Images}

In this section, we compare the performance of eigen images and dynamic images. Both types of images are obtained by pooling RGB values of multiple video frames; the former uses eigen evolution pooling while the latter uses rank pooling.  

\textbf{Globally-pooled and locally-pooled eigen and dynamic images.} We consider both global and local pooling of RGB values, leading to globally-pooled and  locally-pooled images. To compute a global eigen image for a video, we evenly sample 25 frames of a video and apply eigen evolution pooling. To compute the global dynamic image, we use all the frames available.  For locally-pooled eigen and dynamic images, we use a sliding window and apply the corresponding pooling methods to the RGB frames inside the sliding window.  The length of the sliding window is set to 16 frames, which is also the temporal scale used by Dense Trajectories~\cite{Wang-Schmid-ICCV13} and C3D~\cite{Tran-et-al-ICCV15}. 

\textbf{Feature computation and classification.} 
Both eigen images and dynamic images have the same pipeline for feature extraction and classification. 
To recognize the human action in a pooled image (either eigen or dynamic), we  use  a CNN that has been appropriately finetuned,  starting from the spatial-stream VGG-16 model~\cite{wang2016temporal}.  We use the same data augmentation techniques as in \cite{wang2016temporal}, including random cropping and horizontal flipping. We use a dropout ratio of 0.8 at the fc-6 and fc-7 layers. We run the finetuning process for 50 epochs. The learning rate starts at $10^{-3}$ and gradually decreases after every epoch. For evaluation, we use the CNN's output at the final linear layer as the prediction score vector for all the actions.

\begin{table}[t]
\begin{center}
\begin{tabular}{lcc}
\toprule
  & Global Pooling & Local Pooling \\
\midrule
Dynamic Image & 67.2 & 74.9 \\
Eigen Image$_1$ &  68.8 & 77.2 \\
Eigen Image$_2$ &  64.8 & 77.0 \\
Eigen Image$_3$ &  65.5 & 78.1 \\
Eigen Image$_{2+3}$ &  67.8 &  79.0\\
Eigen Image$_{1+2+3}$ &  \textbf{72.7} & \textbf{82.3} \\
\bottomrule
\end{tabular}
\end{center}
\caption{\textbf{Action recognition performance based on temporally pooled RGB images on UCF101 (split1).} We compute eigen images and dynamic images globally for an entire video or locally within a sliding window. We use a finetuned VGG16 model to compute a feature vector representation for each temporally pooled image. }
\label{tab:ei-vs-di}
\end{table}

\textbf{Experimental results.} Table \ref{tab:ei-vs-di} shows the performance of multiple pooling methods. The second and third columns show the results obtained by global pooling and local pooling respectively. As can be seen, local pooling yields better performance than global pooling. This is because: i) locally-pooled images can capture the dynamics of human actions at a finer scale than the globally-pooled images can; and ii) each video lead to many more locally-pooled images than globally-pooled images, so there is more training data to train the feature extraction network for locally-pooled images. 
For both globally-pooled and locally-pooled images, eigen pooling outperforms rank pooling by a wide margin. Eigen Image$_1$ already outperforms Dynamic Image. The combination of Eigen Images 2 and 3 also outperforms  Dynamic Image. The best result is obtained when Eigen Images 1, 2, and 3 are combined.  Notably, the result of Dynamic Image reported here is higher than the result reported in \cite{bilen2016dynamic}, because we use VGG16 model instead of the outdated  AlexNet architecture used by \cite{bilen2016dynamic}.

\subsection{Eigen TSN: pooling of deep-learning features }

In the previous section, we have demonstrated the effectiveness of eigen images, i.e., eigen evolution pooling performed directly on RGB frames. In this section, we show the performance of eigen evolution pooling on deep-learning features. We obtain state-of-the-art results on multiple datasets.

\textbf{TSN features.} 
We use the two-stream Inception-BN model \cite{wang2016temporal} to extract frame-level deep-learning features. The spatial stream of this model inputs an RGB frame  of size $224\times 224 \times 3$ and produces a 1024-dimensional feature vector at the `global ap' layer (after ReLU). The temporal stream is similar with one difference: the input is a stack of 5 consecutive optical flow maps ($224\times 224 \times 10$).

For each video clip, we evenly sample 25 temporal locations. At each location,  the RGB frame (or the stack of optical flows) is  resized to have spatial dimensions of $256 \times 340$ pixels. We extract the feature vectors from five $224\times 224$ regions of the image, the center region and four corners.  We also flip the regions horizontally and compute a feature vector for each flipped region. Subsequently we average the 10 feature vectors and perform $L_2$ normalization to get a single 1024-dimensional feature for each temporal location. Finally, each video is associated with two sequences of 25 1024-dimensional feature vectors, one sequence for the spatial stream and one for the temporal stream.

\textbf{Eigen TSN.} Table \ref{tab:eig-tsn} compares the performance of eigen evolution pooling with average, max, and rank pooling. Each pooling method maps a sequence of feature vectors to a single aggregated feature vector of 1024 dimensions. These aggregated feature vectors can be individually used for action recognition, or they can be combined by concatenation. We perform $L_2$ normalization to the spatial and temporal streams separately, and the feature vectors from the two streams can also be combined. Finally we compute the $l_1$ kernel to train one-vs-all SVMs~\cite{Vapnik-98} for action classification. After learning the classifiers, we use softmax normalization to compute the probability of each action, and evaluate on UCF101 and Hollywood2 datasets.

Table \ref{tab:eig-tsn} shows the performance of different pooling methods on UCF101 and Hollywood2 datasets. On both datasets, the rank pooling is outperformed by average pooling and max pooling. The max pooling is especially effective on UCF101, outperforming average pooling. On Hollywood2, with EEP$_{1+2+3}$, we are able to achieve significant improvement upon the combination among average, rank, and max pooling methods (from 71.1 to 75.0). To achieve the best performance on both datasets, we propose Eigen TSN features, combination between EEP$_{1+2+3}$ evolution pooling and max pooling.

\begin{table}[t]
\begin{center}
\begin{tabular}{lcc}
\toprule
Feature Pooling & UCF101 Split 1 & Hollywood2 \\
\midrule
\small{Mean}~\cite{wang2016temporal} & 94.0 & 66.8 \\
\small{Rank} \cite{bilen2016dynamic} & 91.8 & 54.8 \\
\small{Max}  & \textbf{94.4} & 63.8 \\
\small{EEP$_{1+2+3}$} & \textbf{94.4} & \textbf{75.0} \\
\midrule
\small{Mean + Rank + Max} & \textbf{94.8} & 71.1 \\
\small{EEP$_{1+2+3}$ + Max} & 94.6 & \textbf{75.5} \\
\bottomrule
\end{tabular}
\end{center}
\caption{{\bf Action recognition performance of TSN features using different pooling methods.} The rank pooling is outperformed by average pooling and max pooling. On Hollywood2, EEP$_{1+2+3}$ achieve significant improvement upon the combination among average, rank, and max pooling methods (from 71.1 to 75.0). To achieve the best performance, we propose Eigen TSN, combination between eigen$_{1+2+3}$ evolution pooling and max pooling.}
\label{tab:eig-tsn}
\end{table}

\begin{table}[t]
\begin{center}
\begin{tabular}{lccccccc}
\toprule
\multirow{2}{*}{} & \multicolumn{2}{c}{Eigen Images} & Eigen TSN 
\\ \cmidrule{2-3}
  & Global Pooling   & Local Pooling &  \\
\midrule
Exact eigen evolution pooling &  72.7 &  82.3 & 94.6\\
Approximate pooling by DCT &  72.0 & 82.5 & 94.6\\
\bottomrule
\end{tabular}
\end{center}
\caption{\textbf{ Comparison between the exact and approximate eigen evolution pooling on UCF101(split1).} The exact Eigen Evolution Pooling and its approximation by Discrete Cosine Transform perform similarly for all three methods considered in this work. Compared to EEP, DCT can be directly applied to video representations of variable temporal lengths, without the need to perform fixed-length temporal sampling first. }
\label{tab:eep-vs-dct}
\end{table}

\textbf{Exact versus approximate eigen evolution pooling.}
As aforementioned, the eigen evolution pooling for strongly correlated Markov process, which is often the case, can be approximated using the Discrete Cosine Transform. In Table \ref{tab:eep-vs-dct}, we compare the action recognition performance between exact and approximate eigen evolution pooling. As can be observed, the exact eigen evolution pooling and its approximation by DCT achieve similar action recognition performance for all three methods considered on UCF101 (split1). One advantage of DCT over exact eigen evolution pooling is that DCT can be directly applied to sequential data of any duration, without the need to sample a fixed number of temporal locations.

\textbf{ Comparison to state-of-the-art. }
Table \ref{tab:state-of-the-art} compares our results with the state-of-the-art methods in the last 4 years. 
In particular, one of the most popular action recognition methods is Dense Trajectory Descriptors (DTD)~\cite{Wang-Schmid-ICCV13}, which remain competitive even in the recent surge of deep-learning approach~\cite{Simonyan-Zisserman-NIPS14, Tran-et-al-ICCV15, wang2015action, feichtenhofer2016spatiotemporal}. In fact, most recent state-of-the-art methods~\cite{wang2015action, Wang-Hoai-CVPR16, bilen2016dynamic, fernando2016discriminative, lev2015rnn} propose to combine with Dense Trajectory Descriptors to obtain the best results.
As shown in Table \ref{tab:state-of-the-art}, only using eigen evolution pooling with Temporal Segment Networks, we are able to perform better than or comparably to the previous state-of-the-art methods. 
Combining Eigen TSN and Dense Trajectory-based methods  (DTD~\cite{Wang-Schmid-ICCV13} or VideoDarwin~\cite{Fernando-etal-CVPR15}) significantly advance the state-of-the-art results on both datasets.

\begin{table}[t]
\begin{center}
\begin{tabular}{lrr}
\toprule
Method & UCF101 & Hollywood2 \\
\midrule
Wang and Schmid, 2013 \cite{Wang-Schmid-ICCV13} & $^*$85.5 & 64.7 \\
Hoai and Zisserman, 2014 \cite{Hoai-Zisserman-ACCV14b} & - & 73.6   \\
Simonyan and Zisserman, 2014 \cite{Simonyan-Zisserman-NIPS14} & 88.0 & - \\
Fernando \etal., 2015 \cite{Fernando-etal-CVPR15} & $^*$85.9 & 73.7  \\
Lan \etal., 2015 \cite{lan2015multiskip} & 89.1 & - \\
Tran \etal., 2015 \cite{Tran-et-al-ICCV15} & 90.4 & - \\
Wang \etal., 2015 \cite{wang2015action} & 91.5 & $^*$71.9 \\
Lev \etal., 2015 \cite{lev2015rnn} & 94.1  & - \\
Wang and Hoai, 2016 \cite{Wang-Hoai-CVPR16} &  - & 71.0 \\
Bilen \etal., 2016 \cite{bilen2016dynamic} & 89.1 & - \\
Fernando \etal., 2016\cite{fernando2016discriminative} & 91.4 & 76.7 \\
Wang \etal., 2016 \cite{wang2016temporal} & 94.2 & $^*$66.8 \\
Feichtenhofer \etal., 2016 \cite{feichtenhofer2016spatiotemporal} & 94.6 & - \\
Cherian \etal., 2017 \cite{cherian2017generalized} & 92.3 & - \\
\midrule
Eigen TSN  & 95.3 & 75.5 \\
Eigen TSN + DTD  & \textbf{95.8} & 79.3 \\
Eigen TSN + VideoDarwin  & 95.6 & \textbf{80.5} \\
\bottomrule
\end{tabular}
\end{center}
\caption{{\bf Comparison with state-of-the-art methods on UCF101 and Hollywood2 datasets.} Using eigen evolution pooling with Temporal Segment Networks, combining with Dense Trajectory-based methods (DTD~\cite{Wang-Schmid-ICCV13} or VideoDarwin~\cite{Fernando-etal-CVPR15}), we are able to significantly advance the state-of-the-art performance on both datasets. $^*$ indicates the results obtained by our own re-implementation.}
\label{tab:state-of-the-art}
\end{table}

\section{ Conclusions } \label{sec:conclusion}
We have described eigen evolution pooling, an efficient method  
to compute  compact feature representations for a sequence  of feature vectors. Eigen evolution pooling provides an effective way to capture the long-term and complex dynamics of human actions in video. Eigen evolution pooling can be either used to create eigen images or to aggregate a sequence of CNN features to represent a video. Eigen evolution pooling uses a set of basis functions  to encode the evolution of features over time. The basis functions can be optimally learned from data using PCA or they can be approximated using  the first few basis functions of the Discrete Cosine Transform. We have demonstrated the benefits of eigen evolution pooling over average, max, and rank pooling. Furthermore, we have shown that eigen evolution pooling produces state-of-the-art performance, especially when it is complemented by  Dense Trajectory Descriptors  or VideoDarwin.

{\small
\bibliographystyle{abbrvnat}
\bibliography{shortstrings,pubs2,pubs,egbib}
}

\end{document}